\pdfoutput=1

\documentclass[11pt]{article}

\usepackage[preprint]{acl}

\usepackage{times}
\usepackage{multirow}
\usepackage[table]{xcolor}
\usepackage{latexsym}
\usepackage{amsmath}
\usepackage{amssymb}
\usepackage{hyperref}
\usepackage{subcaption}

\usepackage[T1]{fontenc}

\usepackage[utf8]{inputenc}

\usepackage{inconsolata}

\usepackage{graphicx}

\title{TalkLoRA: Communication-Aware Mixture of Low-Rank Adaptation for Large Language Models}

\author{
  \textbf{Lin Mu\textsuperscript{1}},
  \textbf{Haiyang Wang\textsuperscript{1}},
  \textbf{Li Ni\textsuperscript{1}\footnotemark},
  \textbf{Lei Sang\textsuperscript{1}},
  \textbf{Zhize Wu\textsuperscript{2}},
  \\
  \textbf{Peiquan Jin\textsuperscript{3}},
  \textbf{Yiwen Zhang\textsuperscript{1}}
  \\
  \textsuperscript{1}Anhui University,
  \textsuperscript{2}Hefei University,
  \\
  \textsuperscript{3}University of Science and Technology of China,
\\
  \small{
    \texttt{\{mulin, nili, sanglei, zhangyiwen\}@ahu.edu.cn \{wanghaiyang\}@stu.ahu.edu.cn}
  }
  \\
  \small{
    \texttt{wuzz@hfuu.edu.cn jpq@ustc.edu.cn}
  }
}

\begin{document}
\maketitle

\begin{abstract}
Low-Rank Adaptation (LoRA) enables parameter-efficient fine-tuning of Large Language Models (LLMs), and recent Mixture-of-Experts (MoE) extensions further enhance flexibility by dynamically combining multiple LoRA experts. However, existing MoE-augmented LoRA methods assume that experts operate independently, often leading to unstable routing, expert dominance. In this paper, we propose \textbf{TalkLoRA}, a communication-aware MoELoRA framework that relaxes this independence assumption by introducing expert-level communication prior to routing. TalkLoRA equips low-rank experts with a lightweight \textbf{Talking Module} that enables controlled information exchange across expert subspaces, producing a more robust global signal for routing. Theoretically, we show that expert communication smooths routing dynamics by mitigating perturbation amplification while strictly generalizing existing MoELoRA architectures. Empirically, TalkLoRA consistently outperforms vanilla LoRA and MoELoRA across diverse language understanding and generation tasks, achieving higher parameter efficiency and more balanced expert routing under comparable parameter budgets. These results highlight structured expert communication as a principled and effective enhancement for MoE-based parameter-efficient adaptation. Code is available at \href{https://github.com/why0129/TalkLoRA}{https://github.com/why0129/TalkLoRA}.

\end{abstract}

\renewcommand{\thefootnote}{\fnsymbol{footnote}}
\footnotetext[1]{Corresponding author}

\section{Introduction}

Large language models (LLMs), pre-trained on massive corpora~\cite{llama3,qwen2.5,GPT-4}, have achieved remarkable performance across a wide range of natural language processing tasks~\cite{qin-etal-2023-chatgpt,DDPrompt}. However, adapting such models to domain-specific downstream tasks via full fine-tuning requires updating all parameters, incurring prohibitive memory and computational costs. 
To address this challenge, parameter-efficient fine-tuning (PEFT) methods~\cite{PEFT} such as low-rank
adaptation (LoRA)~\cite{lora} has become a standard approach for adapting LLMs. By injecting low-rank updates into pretrained weights, LoRA enables efficient adaptation while largely preserving the original model’s capabilities. However, standard LoRA employs a single global low-rank update shared across all inputs, which may limit its flexibility under highly heterogeneous prompt and reasoning distributions.~\cite{Mixture-of-LoRAs, modula}.

\begin{figure}[t]
\centering
  \includegraphics[width=\linewidth]{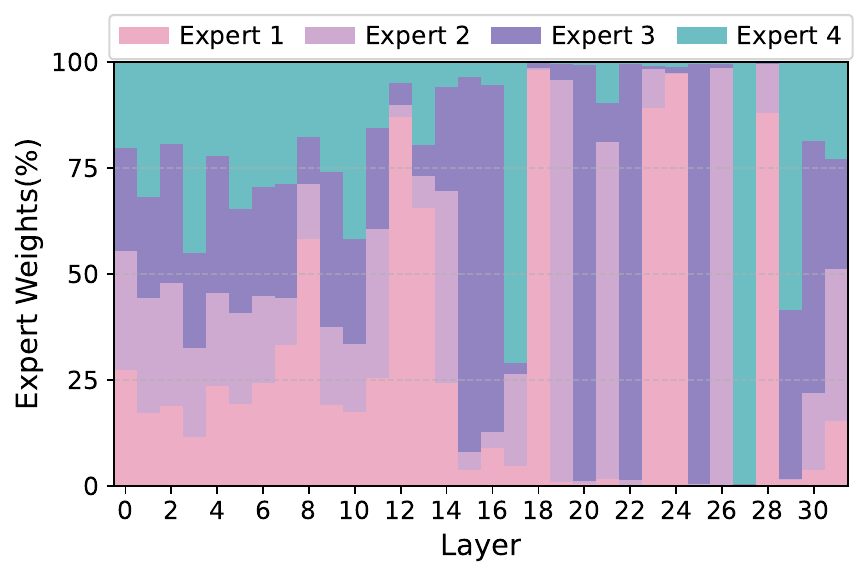} 
  \caption {Average expert weights of different experts during inference in the MoELoRA architecture on the OBQA \cite{obqa} dataset.}
    \label{fig:Motivation}
\end{figure}

A natural extension is to introduce multiple low-rank adapters within a Mixture-of-Experts (MoE) framework~\cite{moe-survey}, where each LoRA module is treated as an expert and a trainable router dynamically combines experts for each input token~\cite{MORE,MOELORA,teamlora}. While such MoE-augmented LoRA (MoELoRA) architectures improve flexibility and enable partial disentanglement of task-shared and task-specific knowledge, they implicitly assume that experts operate independently. In practice, this independence amplifies routing noise, induces sharp and low-entropy gating distributions, and causes the routing mass to concentrate on a small subset of experts~\cite{iclr/Zuo00KHZGZ22}. As illustrated in Figure~\ref{fig:Motivation}, this effect intensifies with network depth: a few experts consistently dominate the routing decisions, while others receive negligible gradient signals, resulting in an ineffective contribution to the model’s overall capacity. Moreover, because experts are trained independently under identical supervision signals and without any explicit coordination mechanism, they tend to learn highly overlapping representations~\cite{LiuD0PCCT24}. This representational redundancy substantially reduces the effective expressivity per parameter, ultimately undermining the goal of parameter-efficient adaptation.

To address these limitations, we draw inspiration from communication mechanisms that relax independence assumptions among model components~\cite{Cross-Stitch, talking-heads}. In particular, talking-head attention~\cite{talking-heads} demonstrated that enabling controlled information exchange among otherwise independent components can significantly improve expressivity and stability. Motivated by this insight, we advocate expert-level communication as a principled extension to MoELoRA. By allowing LoRA experts to exchange compact, task-relevant information during adaptation, such communication facilitates coordination among experts, encourages meaningful specialization while reducing representational redundancy, and smooths parameter updates across expert boundaries.

Building on this idea, we propose \textbf{TalkLoRA}, a communication-aware MoELoRA framework that explicitly relaxes the independence assumption among LoRA experts. TalkLoRA introduces a \textbf{Talking Module} that enables information exchange across low-rank experts. Specifically, we use their internal low-rank projected features as input to the Talking Module, which aggregates global expert information prior to routing. A dense router then assigns weights to all experts and combines their outputs to form the final adaptation added to the frozen pre-trained weights. 
In addition, TalkLoRA incorporates a parameter-sharing mechanism that shares a subset of low-rank factors across layers to reduce the number of trainable parameters. Together, these designs enable TalkLoRA to achieve more expressive and better-balanced routing, resulting in more efficient parameter-efficient adaptation.

Our main contributions can be summarized as follows:

\begin{itemize}
\item We propose TalkLoRA, a communication-aware MoELoRA framework that introduces a lightweight Talking Module to enable information exchange among low-rank experts, effectively relaxing the independence assumption in MoELoRA.
\item We provide a theoretical analysis of TalkLoRA, showing that expert communication strictly enlarges the function class of MoELoRA by enabling cross-expert interactions, and formally demonstrating that such communication promotes more balanced expert routing under mild assumptions.
\item We conduct extensive experiments on multiple datasets and LLMs, confirming the effectiveness of TalkLoRA. In particular, it achieves 87.8\% accuracy on commonsense reasoning with LLaMA3-8B, while demonstrating improved parameter efficiency and more balanced expert routing.
\end{itemize}

\section{Related Works}

\textbf{LoRA and its variants.} LoRA~\cite{lora} freezed the pre-trained weights and injects trainable low-rank decomposition matrices into each transformer layer, effectively approximating weight updates through low-dimensional adaptations ($\Delta W =BA$). This design preserves inference efficiency (as the low-rank matrices can be merged into the original weights during deployment) while maintaining competitive downstream performance. Building on this, numerous variants emerge. For example, DoRA~\cite{dolora} decomposed pre-trained weights into magnitude and direction components, enabling LoRA to focus solely on directional updates. Additionally, DenseLoRA \cite{denselora} compressed LoRA updates into a compact dense matrix to mitigate redundancy in LoRA matrix pairs. HiRA~\cite{hira} addressed this by using a Hadamard product to retain high-rank update parameters, improving the model capacity.

\begin{figure*}[t]
\hspace*{-0.8cm}%
    \begin{minipage}[t]{0.5\linewidth}
		\centering
		\includegraphics[width=1.2\linewidth]{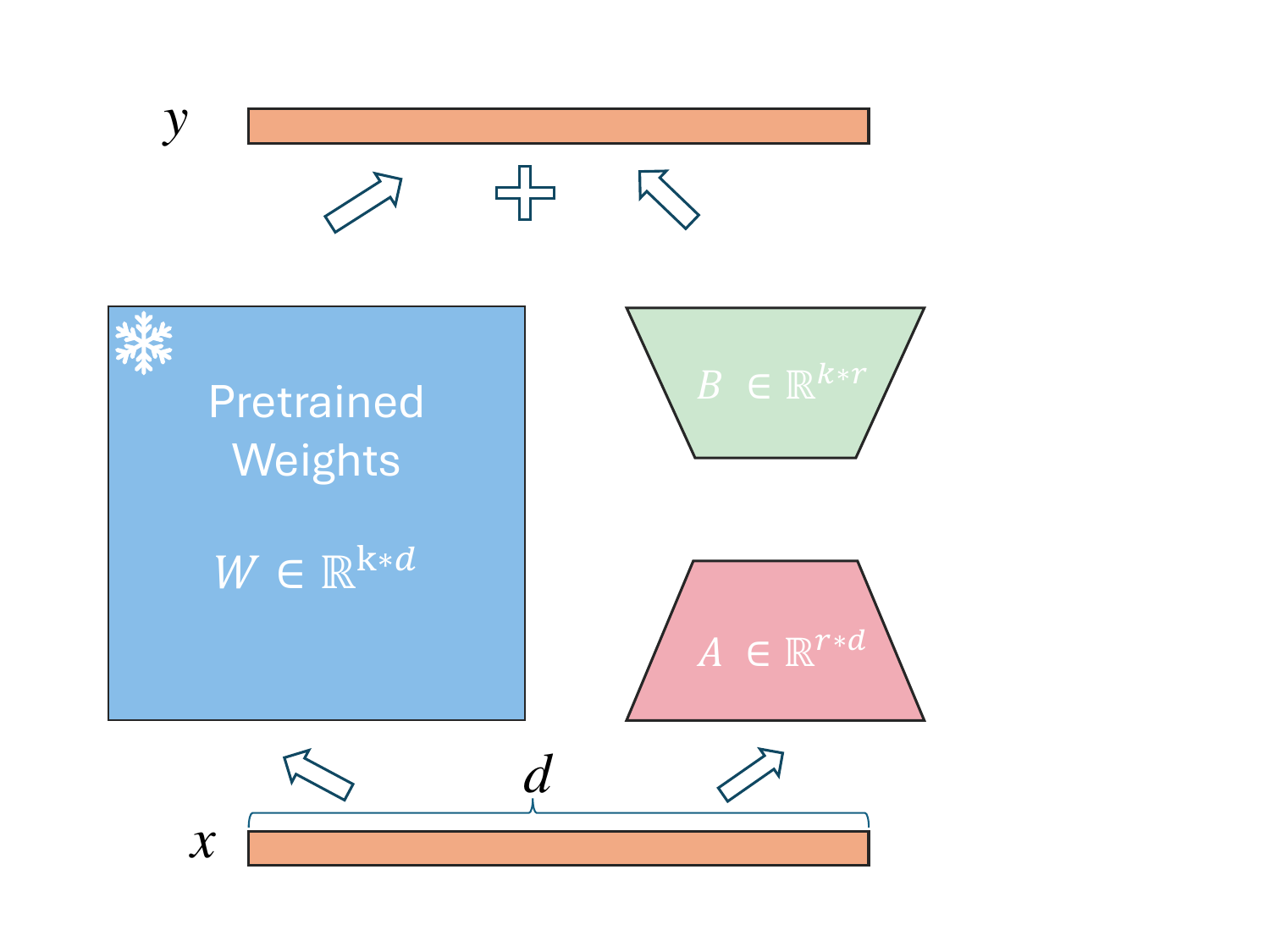}
		\subcaption{LoRA} 
        \label{fig:lora}
    \end{minipage}
    \begin{minipage}[t]{0.5\linewidth}
		\centering
		\includegraphics[width=1.2\linewidth]{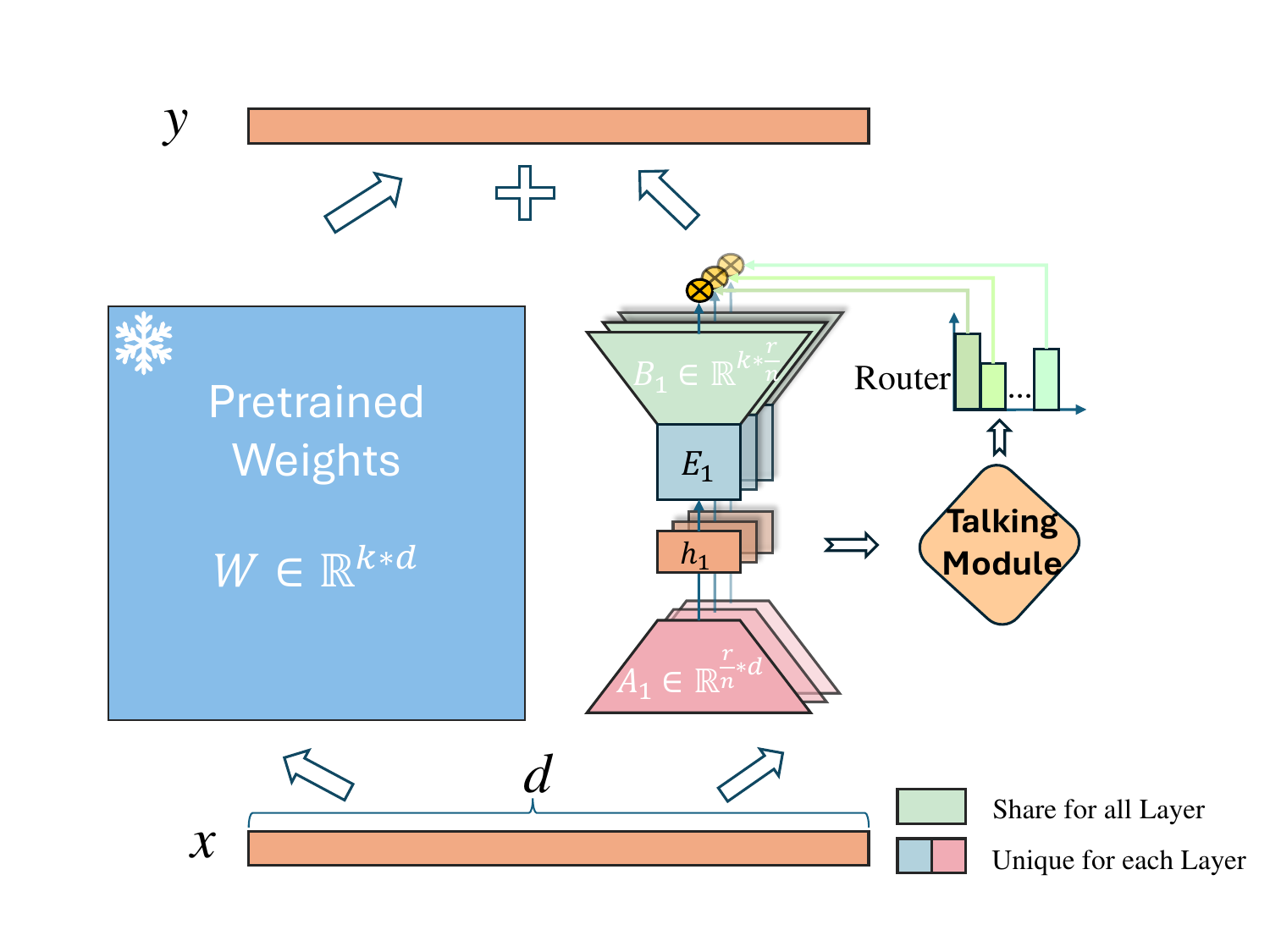}
		\subcaption{TalkLoRA}
	\end{minipage}
\caption{Framework comparison of LoRA(left) and TalkLoRA(right).}
\label{fig:TalkLoRA}
\end{figure*}

\textbf{MoELoRA.} Methods that integrate the Mixture-of-Experts (MoE) paradigm with LoRA have recently garnered considerable research attention. LoRAMoE~\cite{loramoe} freezed the backbone model and forces a portion of LoRAs to focus on leveraging world knowledge to solve downstream tasks, to alleviate world knowledge-edge forgetting. MoSLD's~\cite{mosld} core idea was to share the matrix $A$ as the general-feature matrix and keep matrix $B$ as specific-feature matrix and applies dropout to mitigate the imbalance in parameter updates. TeamLoRA~\cite{teamlora}, consisting of a collaboration and competition module for experts, thus achieving the right balance of effectiveness and efficiency. MTL-LoRA~\cite{mtllora} introduced additional task-adaptive parameters to distinguish task-specific information and captures shared knowledge across tasks in a low-dimensional space.

Unlike existing MoELoRA architectures where experts operate in isolation, TalkLoRA introduces an expert-level communication mechanism. This facilitates inter-expert collaboration and effectively resolves the load imbalance issue.

\section{Methodology}
In this section, we elaborate on the technical details of TalkLoRA. An overview of the proposed architecture is presented in Figure~\ref{fig:TalkLoRA}.
\subsection{Background}
\textbf{Low-Rank Adaptation.}
The core idea of LoRA \cite{lora} is to freeze the original model parameters and inject a low-rank decomposition into the weight updates. Specifically, a pretrained weight matrix $ W_0 \in \mathbb{R}^{k \times d} $ is frozen, and two trainable low-rank matrices $ A \in \mathbb{R}^{r \times d} $ and $ B \in \mathbb{R}^{k \times r} $ in LoRA handle the parameter updates. The rank $ r $ is much smaller than $ d $ and $ k $ (i.e. $r << min(d,k)$). Given an input $ x \in \mathbb{R}^{d} $ to the LLMs, the output $ y \in \mathbb{R}^{k} $ after LoRA is expressed as:
\begin{equation}\label{equ:lora}
    y = (W_0 + \Delta W)x = W_0x +BAx,
\end{equation}
where matrix $A$ undergoes Kaiming initialization~\cite{kaiming}, while matrix $B$ receives zero initialization, ensuring that fine-tuning initially preserves the original output. 
During inference, $ \Delta W $ merges with $ W_0 $ (i.e. $W^{'}=W_0 + \Delta W$), eliminating additional latency in the adapted model.

\textbf{Mixture-of-Experts LoRA (MoELoRA).}
MoE~\cite{6797059,Mixture-of-Experts} models constitute a neural network architecture designed to enhance model capacity and computational efficiency. The core principle activates a subset of expert for a given input or employs dense routing to activate all experts with assigned weights. 

LoRA integrates with MoE by treating the product of matrices $A_i\in \mathbb{R}^{\frac{r}{n} \times d}$ and $B_i\in \mathbb{R}^{k \times \frac{r}{n}}$ as a single expert. Each MoELoRA layer contains $ n $ LoRA experts. The forward process of the layer is expressed as:
 \begin{equation}\label{equ:moelora}
    y = W_0x + \Delta W x = W_0x+ \sum_{i=1}^{n}g_i(x)  B_i A_i x.
 \end{equation}
To balance the contribution of these experts, MoELoRA use a gate function $g$, which acts as a router network. This network is a fully connected layer with trainable weights $W_g \in \mathbb{R}^{n \times d}$. It is followed by a $softmax$ function that takes a $x$ as input. 
\begin{equation}\label{equ:moerouter}
    g(x) = softmax(W_g x).
\end{equation}

\subsection{TalkLoRA Architecture}
To address experts operating in isolation and the load imbalance in traditional MoELoRA architectures. TalkLoRA incorporates a Talking Module to enable information exchange among experts and direct the assignment of routing weights. Additionally, inspired by DenseLoRA~\cite{denselora}, we adopt a parameter sharing strategy for a subset of trainable parameters to reduce redundancy and refine experts. The following details the integration of TalkLoRA into LLMs.

\textbf{Expert Component:}
We decompose the original LoRA into $n$ sub-LoRA experts. We further parameterize the up-projection matrix of each LoRA expert as the product of matrix $E_i \in \mathbb{R}^{\frac{r}{n} \times \frac{r}{n}}$ and matrix $B_i \in \mathbb{R}^{k \times \frac{r}{n}}$. Each $A_i \in \mathbb{R}^{\frac{r}{n} \times d}$ and $E_i$ learns domain-specific knowledge. Subsequently, $B_i$ restores the expert dimension to match the original weight matrix output dimension, ensuring compatibility with the LLMs. We use
$x\in \mathbb{R}^{d}$ as the input to each expert, and $y_i\in \mathbb{R}^{k}$ denotes the corresponding expert output. This process is formulated as:
\begin{equation}\label{equ:expert}
    y_i = B_i E_i A_ix,
\end{equation}
where $i$ ranges from 1 to $n$. Here, $r$ denotes the total rank of TalkLoRA, and $n$ represents the number of experts.

\textbf{Talking Module:}
This module guides the router in weight allocation and relaxes the independence assumption among experts. It enables information exchange among experts prior to routing. Formally, we define:
\begin{equation}\label{equ:talking}
    \tilde{h_i} = \sum_{j=1}^{n} C_{ij} h_j,
\end{equation}
where $C \in \mathbb{R}^{n \times n}$ is a learnable communication matrix and $h_j=A_jx \in \mathbb{R}^\frac{r}{n}$ serves as the internal representation of expert $j$.

This operation allows each expert to integrate compact, task-relevant signals from other experts while preserving its own specialization. The Talking Module is lightweight, adding only $O(n^2)$ parameters.

\textbf{Routing:}
Unlike traditional routing, which relies solely on the original input $x$ for decision-making, TalkLoRA performs routing decisions based on the communicated representations $\tilde{h} \in \mathbb{R}^\frac{r}{n}$. The process is formulated as follows:
\begin{equation}\label{equ:talkrouter}
    g([\tilde{h_1}, \tilde{h_2},...,\tilde{h_n}]) = softmax(W_g [\tilde{h_1}, \tilde{h_2},...,\tilde{h_n}]),
\end{equation}
where $W_g \in \mathbb{R}^{n \times r} $ denotes the router parameters. By conditioning routing on globally informed expert features, TalkLoRA mitigates routing overconfidence and reduces sensitivity to local noise.

The overall adaptation process in TalkLoRA can be mathematically formulated as follows, combining the frozen pre-trained weights $W_0\in\mathbb{R}^{k\times d}$ with the improved experts and router:
 \begin{equation}\label{equ:talklora}
 \begin{aligned}
    y &= W_0x +  \Delta W x \\
      &= W_0x + \sum_{i=1}^{n}g([\tilde{h_1}, \tilde{h_2},...,\tilde{h_n}]) y_i.
      \end{aligned}
 \end{equation}

\begin{table*}[t]
    \begin{center} 
    \renewcommand{\arraystretch}{1.0}
    \setlength{\tabcolsep}{2pt}
    \resizebox{\linewidth}{!}{
    \begin{tabular}{llcccccccccc}
        \hline 
        \textbf{LLM} & \textbf{Method} & \textbf{\#Param(\%)} & \textbf{BoolQ} & \textbf{PIQA} &  \textbf{SIQA} & \textbf{ARC-c} & \textbf{ARC-e} & \textbf{OBQA} & \textbf{HellaS.} &  \textbf{WinoG.} & \textbf{Avg.}\\ \hline
        \cellcolor{white}\multirow{4}*{Qwen2.5-7B} 
        & LoRA$^{\dagger}$ ($r=16$)  & 0.4 & 60.0 & 73.6 & 70.0 & 71.7 & 85.9 & 74.4 & 78.6 & 75.8 & 73.8 \\
        & HiRA$^{\dagger}$ ($r=16$)  & 0.4 & 69.0 & 88.3 & 80.8 & 88.7 & 95.4 & 88.0 & 92.3 & 81.0 & 85.4 \\
        & TeamLoRA$^{\dagger}$ ($r=16$)    & 0.4  &74.6  &90.0  &82.3  &88.5  &95.9  &92.2  &95.4  &89.0  &88.5  \\
        & \cellcolor{gray!20}TalkLoRA ($r=16$)   & \cellcolor{gray!20}\textbf{0.2}
        & \cellcolor{gray!20}73.6 & \cellcolor{gray!20}90.9 & \cellcolor{gray!20}83.0
        & \cellcolor{gray!20}89.6 & \cellcolor{gray!20}96.5 & \cellcolor{gray!20}92.8
        & \cellcolor{gray!20}95.5 & \cellcolor{gray!20}89.9 & \cellcolor{gray!20}89.0 \\ \hline
        \cellcolor{white}\multirow{7}*{LLaMA2-7B} 
        & LoRA ($r=32$)  & 0.8 & 69.8 & 79.9 & 79.5 & 64.7 & 79.8 & 81.0 & 83.6 & 82.6 & 77.6 \\
        & DoRA ($r=32$)  & 0.8 & 71.8 & 83.7 & 76.0 & 68.2 & 83.7 & 82.4 & 89.1 & 82.6 & 79.7 \\
        & HiRA ($r=32$)  & 0.8 & 71.2 & 83.4 & 79.5 & 73.8 & 86.7 & 84.6 & 88.1 & 84.0 & 81.4 \\
        & MixLoRA ($r=16$)     & 2.9 & 72.7 & 83.2 & 78.0 & 58.1 & 77.7 & 81.6 & 93.1 & 76.8 & 77.6 \\
        & MoELoRA ($r=16$)     & 0.3 & 68.0 & 83.5 & 70.4 & 61.5 & 86.8 & 83.2 & 90.6 & 82.5 & 78.3 \\
        & TeamLoRA$^{\dagger}$ ($r=32$)     & 0.9 & 70.6 & 82.8 & 79.0 & 72.4 & 86.0 & 81.4 & 86.2 & 83.3 & 80.2 \\
        & \cellcolor{gray!20}TalkLoRA ($r=16$) & \cellcolor{gray!20}\textbf{0.2}
        & \cellcolor{gray!20}72.6 & \cellcolor{gray!20}83.9 & \cellcolor{gray!20}81.5
        & \cellcolor{gray!20}73.5 & \cellcolor{gray!20}88.0 & \cellcolor{gray!20}86.0
        & \cellcolor{gray!20}89.5 & \cellcolor{gray!20}85.2 & \cellcolor{gray!20}82.5\\
    
        & \cellcolor{gray!20}TalkLoRA ($r=32$) & \cellcolor{gray!20}\textbf{0.4}
        & \cellcolor{gray!20}73.1 & \cellcolor{gray!20}84.8 & \cellcolor{gray!20}80.9
        & \cellcolor{gray!20}75.7 & \cellcolor{gray!20}87.9 & \cellcolor{gray!20}84.8
        & \cellcolor{gray!20}89.2 & \cellcolor{gray!20}86.5 & \cellcolor{gray!20}82.9 \\ \hline
        \multirow{9}*{LLaMA3-8B}
        & LoRA ($r=32$)   & 0.7 & 70.8 & 85.2 & 79.9 & 71.2 & 84.2 & 79.0 & 91.7 & 84.3 & 80.8 \\
        & DoRA ($r=32$)   & 0.7 & 74.6 & 89.3 & 79.9 & 80.4 & 90.5 & 85.8 & 95.5 & 85.6 & 85.2 \\
        & HiRA ($r=32$)   & 0.7 & 75.4 & 89.7 & 81.2 & 82.9 & 93.3 & 88.3 & 95.4 & 87.7 & 86.7 \\
        & MixLoRA ($r=16$)& 3.0 & 75.0 & 87.6 & 78.8 & 79.9 & 86.5 & 84.8 & 93.3 & 82.1 & 83.5 \\
        & MoELoRA$^{\dagger}$ ($r=32$)& 0.7 & 74,6 & 89.1 & 82.3 & 82.8 & 92.7 & 87.6 & 95.3 & 88.5 & 86.6 \\
        & TeamLoRA$^{\dagger}$ ($r=32$)& 0.7 & 74.3 & 88.2 & 81.8 & 81.7 & 92.8 & 88.0 & 95.4 & 89.0 & 86.4 \\
        & \cellcolor{gray!20}TalkLoRA ($r=16$)& \cellcolor{gray!20}\textbf{0.2}
        & \cellcolor{gray!20}75.3 & \cellcolor{gray!20}88.8 & \cellcolor{gray!20}82.3
        & \cellcolor{gray!20}84.3 & \cellcolor{gray!20}93.2 & \cellcolor{gray!20}89.2
        & \cellcolor{gray!20}96.2 &	\cellcolor{gray!20}89.6 & \cellcolor{gray!20}87.4\\ 
        
        & \cellcolor{gray!20}TalkLoRA ($r=32$) & \cellcolor{gray!20}\textbf{0.4}
        & \cellcolor{gray!20}76.1 & \cellcolor{gray!20}89.6	& \cellcolor{gray!20}82.3
        & \cellcolor{gray!20}84.5 & \cellcolor{gray!20}93.9 & \cellcolor{gray!20}89.4
        & \cellcolor{gray!20}96.0 & \cellcolor{gray!20}89.2 & \cellcolor{gray!20}87.6\\
        
        & \cellcolor{gray!20}TalkLoRA$^{\ast}$ ($r=32$)& \cellcolor{gray!20}\textbf{0.4}
        & \cellcolor{gray!20}75.1 & \cellcolor{gray!20}89.2 & \cellcolor{gray!20}83.6
        & \cellcolor{gray!20}84.9 & \cellcolor{gray!20}93.9	& \cellcolor{gray!20}88.8
        & \cellcolor{gray!20}96.4 & \cellcolor{gray!20}90.1 & \cellcolor{gray!20}87.8\\
        \hline  
    \end{tabular}}
    \end{center}
    \caption{Accuracy(\%) comparison of various methods fine-tuning Qwen2.5-7B, LLaMA2-7B and LLaMA3-8B on the commonsense reasoning tasks. Results for LoRA, DoRA, and HiRA are sourced from~\cite{hira}, MixLoRA and MoELoRA are sourced from \cite{mixlora} and \cite{mtllora}, respectively. $^{\dagger}$ indicates results reproduced using the same configuration as TalkLoRA. $^{\ast}$ denotes the use of 8 experts.}
    \label{tab:Commonsense Reasoning}
\end{table*}

\textbf{Initialization Strategies:}
Similar to LoRA's initialization strategy, we apply Kaiming initialization~\cite{kaiming} to matrices $A_i\in \mathbb{R}^{\frac{r}{n} \times d}$ and $E_i\in \mathbb{R}^{\frac{r}{n} \times \frac{r}{n}}$ in the expert component of TalkLoRA, and zero initialization to $B_i\in \mathbb{R}^{k\times \frac{r}{n}}$, ensuring that initial training preserves the original output. For the routing component, matrices $C\in \mathbb{R}^{n \times n}$ and $W_g\in\mathbb{R}^{n \times r}$ also receive Kaiming initialization.

Notably, $B_i$ is shared across different adaptation layers, whereas $A_i$ and $E_i$ remain unique to each layer. This strategy substantially reduces computational cost while preserving overall performance.

\subsection{Analysis of TalkLoRA}

\textbf{Expressive Power Analysis.} 
In MoELoRA, each expert operates within an independent low-rank subspace, which limits the expressivity of the resulting parameter updates to isolated expert contributions. By contrast, TalkLoRA introduces expert-level communication that enables linear interactions across experts prior to routing, thereby allowing parameter updates to span mixed expert subspaces.

\begin{itemize}
    \item When $C_{ij}=0$ for all $i \neq j$, TalkLoRA degenerates  to MoELoRA, where experts remain fully independent.. 
    \item When $C_{ij} \neq 0$ for some $i\neq j$, the representation of expert $h_i$ can incorporate information from expert $h_j$, resulting in cross-expert interactions that are inaccessible to MoELoRA.
\end{itemize}
 Since MoELoRA is recovered as a degenerate case of TalkLoRA when expert communication vanishes, TalkLoRA strictly subsumes the function class of MoELoRA, yielding strictly greater expressive power.

\textbf{Routing Sensitivity Analysis:} 
Let $\tilde{h} = (C \otimes I) h$ denote the communicated expert representations used as input to the router. When the expert communication matrix $C$ is non-expansive (the experimental verification shown in the Appendix ~\ref{sec:add experimental}.), perturbations in the input are not amplified through the communication module, resulting in bounded changes in the router input. 

Specifically, an input perturbation is first projected into the low-rank adaptation space and distributed across experts. The communication module then aggregates expert representations via a linear transformation. If this transformation is non-expansive, it smooths local variations by sharing information across experts rather than amplifying noise. Consequently, the router operates on more stable and less noisy representations, leading to smoother and more robust expert selection.

In contrast, MoELoRA corresponds to the degenerate case without expert communication, where routing decisions depend solely on isolated expert signals and are therefore more sensitive to input perturbations. A formal analysis under standard boundedness assumptions is provided in Appendix~\ref{sec:analysis}.

\begin{table*}[t]
    \begin{center} 
    \renewcommand{\arraystretch}{1.0}
    \setlength{\tabcolsep}{8pt}
    \resizebox{\linewidth}{!}{
    \begin{tabular}{l c c c c c c c c}
    \hline
    \textbf{Method}  & \textbf{\#Param} & \textbf{SST-2} & \textbf{MRPC} 
            & \textbf{CoLA} & \textbf{QNLI} & \textbf{RTE} & \textbf{STS-B} & \textbf{Avg} \\ 
    \hline
    BitFit  & 0.1M 
            & 94.0$_{\pm 0.87}$ & 88.1$_{\pm 1.57}$ & 54.0$_{\pm 3.07}$ 
            & 91.0$_{\pm 0.05}$ & 69.8$_{\pm 1.51}$ & 89.5$_{\pm 0.35}$ & 81.1 \\
    IA3     & 0.06M 
            & 93.4$_{\pm 0.00}$ & 86.4$_{\pm 0.00}$ & 57.8$_{\pm 0.00}$ 
            & 91.1$_{\pm 0.00}$ & 73.5$_{\pm 0.00}$ & 88.5$_{\pm 0.00}$ & 81.8 \\
    LoReFT  & 0.02M 
            & 93.4$_{\pm 0.64}$ & 89.2$_{\pm 2.62}$ & 60.4$_{\pm 2.60}$ 
            & 91.2$_{\pm 0.25}$ & \textbf{79.0}$_{\pm 1.76}$ & 90.0$_{\pm 0.29}$ & 83.9 \\
    RED     & 0.02M 
            & 93.9$_{\pm 0.31}$ & 89.2$_{\pm 0.98}$ & 61.0$_{\pm 2.96}$ 
            & 90.7$_{\pm 0.35}$ & 78.0$_{\pm 2.06}$ & 90.4$_{\pm 0.32}$ & 83.9 \\
    LoRA    & 0.3M 
            & 93.9$_{\pm 0.49}$ & 88.7$_{\pm 0.76}$ & 59.7$_{\pm 4.36}$ 
            & 92.6$_{\pm 0.10}$ & 75.3$_{\pm 2.79}$ & 90.3$_{\pm 0.54}$ & 83.4 \\
    Adapter & 0.4M 
            & 93.3$_{\pm 0.40}$ & 88.4$_{\pm 1.54}$ & 60.9$_{\pm 3.09}$ 
            & 92.5$_{\pm 0.02}$ & 76.5$_{\pm 2.26}$ & 90.5$_{\pm 0.35}$ & 83.7 \\
    DeLoRA  & 0.3M 
            & 94.1$_{\pm 0.70}$ & 89.0$_{\pm 0.96}$ & 63.6$_{\pm 1.52}$ 
            & 92.8$_{\pm 0.51}$ & 77.1$_{\pm 3.65}$ & \textbf{90.9}$_{\pm 0.31}$ 
            & 84.6 \\
    \rowcolor{gray!20}TalkLoRA 
            & 0.3M
            & \textbf{94.2}$_{\pm0.37}$  & \textbf{89.3}$_{\pm1.37}$
            & \textbf{64.2}$_{\pm2.51}$  
            & \textbf{93.0}$_{\pm0.32}$  & 77.6$_{\pm0.15}$  & \textbf{90.9}$_{\pm0.52}$
            & \textbf{84.9} \\
    \hline
    \end{tabular}}
    \end{center}
    \caption{Performance comparison of RoBERTa-base model fine-tuned by TalkLoRA and other PEFT baseline methods on the GLUE benchmark (without MNLI and QQP). We report Matthew’s correlation for CoLA,Pearson correlation for STS-B, and accuracy for the remaining tasks. All baseline results are sourced from~\cite{delora}.}
    \label{tab:GLUEresults}
\end{table*}

\section{Experiments}
First, we fine-tune the Qwen2.5-7B~\cite{qwen2.5}, LLaMA2-7B~\cite{llama2} and LLaMA-8B~\cite{llama3} models on commonsense reasoning tasks and compare the performance of TalkLoRA against LoRA and its variants.
Next, we evaluate smaller LLM, specifically RoBERTa-base (125M)~\cite{roberta}. We compare different methods for fine-tuning RoBERTa-base on the General Language Understanding Evaluation (GLUE) benchmark~\cite{glue}.
We then identify the optimal placement of TalkLoRA within the Transformer architecture. 
Subsequently, we conduct a robustness analysis of TalkLoRA, examining its performance variation across different expert ranks and numbers of experts. Finally, we verify the contribution of the core module in TalkLoRA.

\subsection{Commonsense Reasoning}
To demonstrate the performance of TalkLoRA on commonsense reasoning tasks, which include BoolQ~\cite{boolq}, PIQA~\cite{piqa}, SIQA~\cite{siqa}, HellaS.~\cite{hellaswag}, WinoG.~\cite{wino}, ARC-c and ARC-e~\cite{arc}, and OBQA~\cite{obqa}. Detailed descriptions of these datasets appear in the Appendix~\ref{dataset:CR}.

\textbf{Experimental Details.}
For the commonsense reasoning dataset, we fine-tune LLMs for 2 epochs, with evaluation on the validation set every 80 steps. The best checkpoint is selected for the final testing. We use AdamW optimizer~\cite{adamw}, detailed hyperparameter settings appear in the Table~\ref{Hyperparameters:CR}. 
We establish LoRA and its variants as baselines, including DoRA and HiRA, and also compare TalkLoRA with related MoELoRA-based methods: MixLoRA~\cite{mixlora} and TeamLoRA~\cite{teamlora}.
%
All the experiments are conducted using 4 Nvidia 24GB 3090 GPU.

\textbf{Main Results.} Table~\ref{tab:Commonsense Reasoning} presents the commonsense reasoning performance of various parameter-efficient fine-tuning (PEFT) methods on the Qwen2.5-7B, LLaMA2-7B and LLaMA3-8B base models, reporting accuracy across eight standard benchmarks and their average (Avg.). The proposed TalkLoRA comprehensively outperforms most of the baselines under extremely low parameter budgets (0.2\%–0.4\%), demonstrating substantial performance gains.

On Qwen2.5-7B: The most significant observation is that TalkLoRA achieves the highest average accuracy of 89.0\%, surpassing the strong baseline TeamLoRA (88.5\%) and significantly outperforming the standard LoRA (73.8\%).

On LLaMA2-7B: TalkLoRA achieves 82.9\% average accuracy with only 0.4\% trainable parameters, outperforming the DoRA by 3.2\% and standard LoRA by 5.3\%. Even in $r=16$ (0.2\% parameters), it reaches 82.5\%, surpassing the HiRA (81.4\%).

On LLaMA3-8B: TalkLoRA achieves an average accuracy of 87.6\%, surpassing HiRA by 0.9\% and DoRA by 2.4\%. TalkLoRA with $r=16$ attains 87.4\%, outperforming all $r=32$ baselines while doubling parameter efficiency. More remarkably, with total rank $r=32$ and 8 experts—yielding a per-expert rank of only 4 (half the conventional minimum)—it further improves the best result (87.6\%) by 0.2\%.


\subsection{Natural Language Understanding}
To evaluate TalkLoRA on smaller language models, we fine-tune RoBERTa-base~\cite{roberta} and use GLUE~\cite{glue} as the evaluation benchmark. The benchmark comprises eight widely used tasks covering syntactic acceptability (CoLA), sentiment classification (SST-2), paraphrase detection (MRPC, QQP), semantic similarity measurement (STS-B), and natural language inference (MNLI, QNLI, RTE). Due to computational constraints, we exclude the resource-intensive MNLI and QQP tasks. Detailed descriptions of these datasets appear in the Appendix~\ref{dataset:glue}.

\textbf{Experimental Details.}
First, we partition the validation set into two subsets. Detailed dataset sizes appear in the Table \ref{GLUE}. We then construct hyperparameter groups comprising different learning rates, training epochs, and batch sizes. After each training epoch, we evaluate on the first subset; only when the highest validation score is achieved do we assess performance on the second subset. Finally, we apply the hyperparameter configuration yielding the best result across multiple random seeds and report the average to ensure experimental reliability. Consistent with prior work ~\cite{delora}, we apply TalkLoRA to the Query and Value projections, setting the total rank to 16 and the number of experts to 4 to ensure fairness in trainable parameters. We use BitFit~\cite{bitfit}, IA3~\cite{IA3}, LoReFT~\cite{REft}, RED~\cite{RED}, LoRA, Adapter~\cite{PEFT} and DeLoRA~\cite{delora} under the identical experimental protocol as controls. Detailed hyperparameter settings appear in the Table~\ref{Hyperparameters:glue}.

\begin{table}[t]
    \begin{center} 
    \resizebox{\linewidth}{!}{
    \begin{tabular}{cccc}
    \hline 
    \textbf{\#Param(\%)}& \textbf{LoRA} & \textbf{TalkLoRA} & \textbf{Avg.} \\ \hline
    0.74 & QKVUD & -     & 80.8     \\
    0.17 & -     & QKV   & 87.0     \\
    0.24 & -     & UD    & 87.3     \\
    0.41 & -     & QKVUD & \textbf{87.6}\\\hline  
    \end{tabular}}
    \end{center}
    \caption{Accuracy(\%) comparison of TalkLoRA with several different tuning granularity fine-tuning LLaMA3-8B. Each module is represented by its first letter as follows: (Q)uery, (K)ey, (V)alue, (U)p, (D)own.}
    \label{tab:Tuninggra}
\end{table}

\begin{figure}[t]
\centering
  \includegraphics[width=\linewidth]{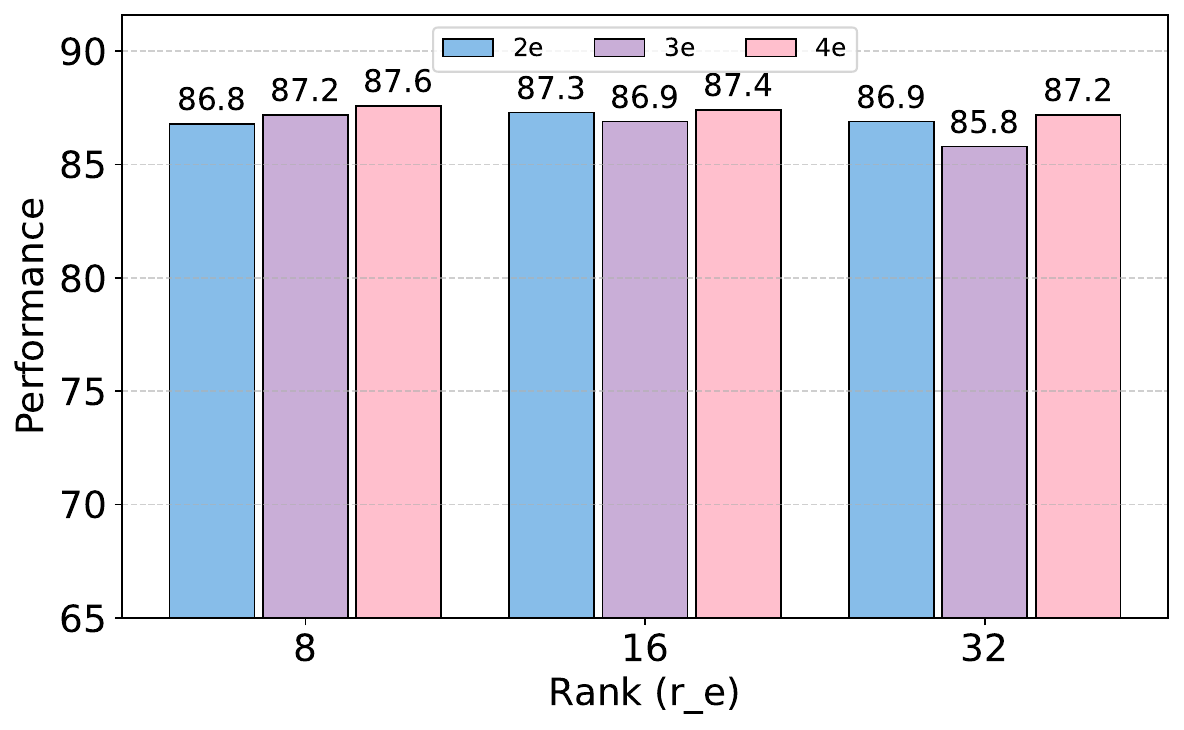} 
  \caption {Robustness analysis of TalkLoRA. "2e, 3e, 4e" denotes different numbers of experts.}
    \label{fig:robustness}
\end{figure}

\textbf{Main Results.} Table~\ref{tab:GLUEresults} presents performance comparisons of TalkLoRA and other PEFT methods on GLUE, reporting the mean and standard deviation across five random seeds for each task. Compared with representative PEFT baselines, TalkLoRA obtains the best results on SST-2, MRPC, CoLA, QNLI, and overall average score, reaching 94.2, 89.3, 64.2, 93.0, and 84.9, respectively. In addition, it delivers competitive performance on the remaining tasks, demonstrating strong generalization capability. Overall, under a comparable parameter budget, TalkLoRA substantially outperforms existing methods and exhibits consistent robustness across tasks.

\begin{table}[t]
    \begin{center}
    \setlength{\tabcolsep}{12pt}
    \resizebox{\linewidth}{!}{
    \begin{tabular}{lcccccc}
        \hline 
        \textbf{Method} & \textbf{\# Param(\%)} &\textbf{Avg.} \\ 
       \hline
        TalkLoRA     &  0.4   & 87.6 \\
        w/o Sharing  &  0.7   & 87.7 \\
        w/o Talking   &  0.4   & 86.5 \\  
        \hline  
    \end{tabular}}
    \end{center}
    \caption{Accuracy (\%) comparison of several variants of TalkLoRA.}
    \label{tab:Variant}
\end{table}

\begin{figure*}[t]
\centering
\vspace*{-1cm}
  \includegraphics[width=\linewidth]{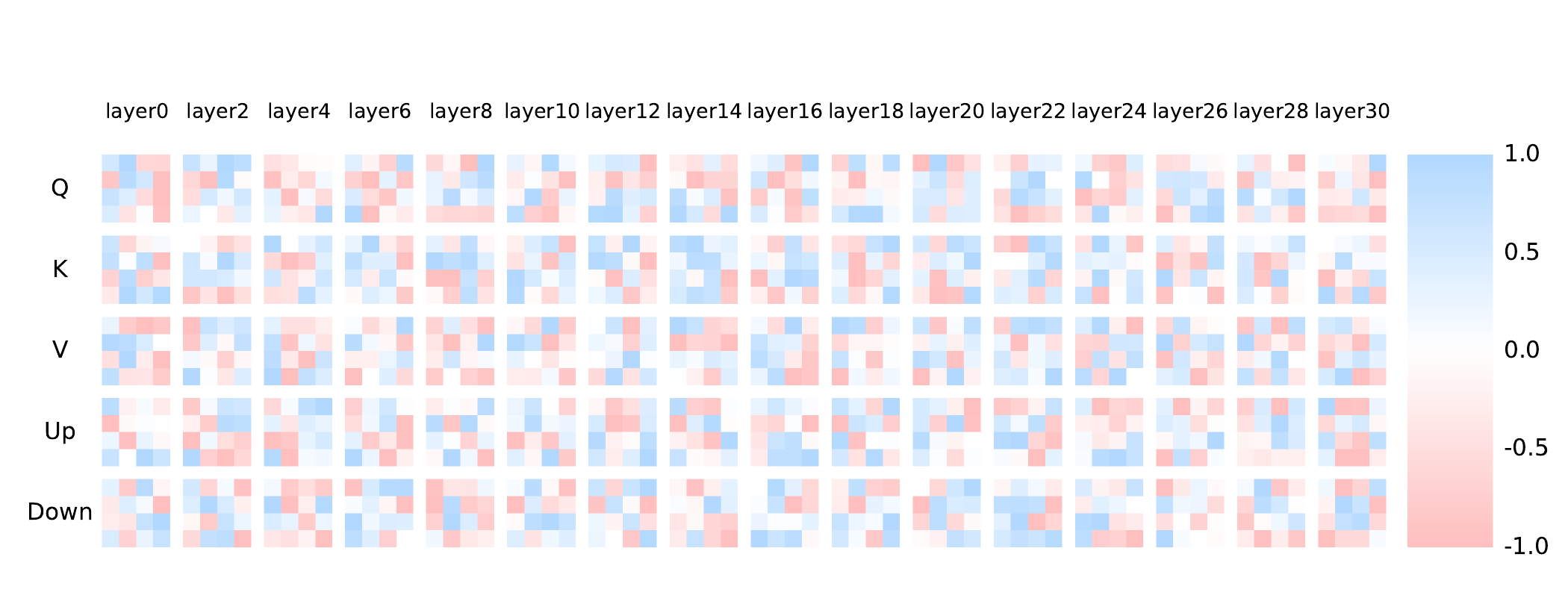} 
  \caption {Visualization of the learned communication matrix $C$. All entries in each matrix are normalized to [-1, 1]. Due to space limitations, we select only even-numbered layers for visualization.}
    \label{fig:talk}
\end{figure*}

\begin{figure*}[t]                 
\centering                         
\begin{minipage}[t]{0.48\linewidth} 
    \centering
    \includegraphics[width=\linewidth]{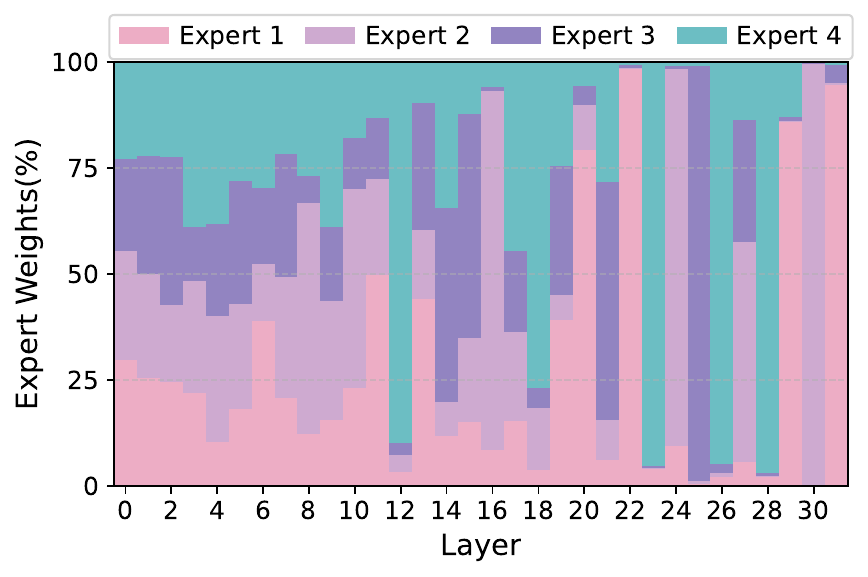}
    \subcaption{Without Talking Module}
    \label{fig:Without Talking Module}
\end{minipage}
\hfill
\begin{minipage}[t]{0.48\linewidth}
    \centering
    \includegraphics[width=\linewidth]{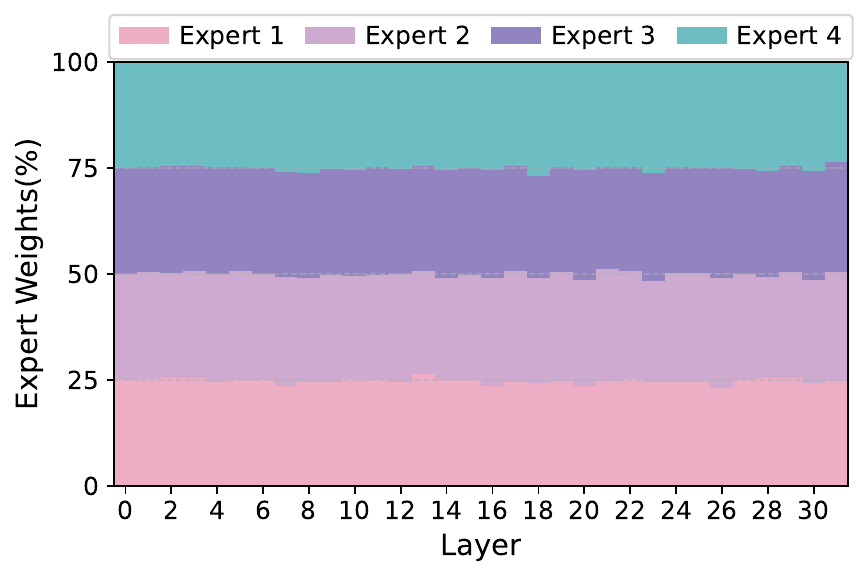}
    \subcaption{TalkLoRA}
    \label{fig:With Talking Module}
\end{minipage}

\caption{Router load visualization of without Talking Module (left) and TalkLoRA (right).}
\label{fig:router}                
\end{figure*}

\subsection{Tuning Granularity Analysis} In this section, we analyze the effects of applying TalkLoRA to different adaptation layers in LLMs. We specifically investigate performance in the self-attention module and the feed-forward network, targeting the (Q)uery, (K)ey, (V)alue, (U)p, and (D)own weight matrices. We fine-tune LLaMA3-8B on commonsense reasoning tasks with rank 32 and 4 experts. The result, shown in Table~\ref{tab:Tuninggra}, highlight several key observations:

When tuning only the QKV (0.17\%) or UD (0.24\%), TalkLoRA achieves average accuracies above 87.0\%, substantially higher than standard LoRA’s 80.8\% accuracy with a much larger parameter budget of 0.74\%. This indicates that TalkLoRA is significantly more parameter-efficient. Furthermore, when applying the full QKVUD configuration, TalkLoRA reaches the highest accuracy of 87.6\% with only 0.41\% parameters, demonstrating both superior performance and compact parameter usage. Overall, TalkLoRA delivers more stable and effective improvements across different tuning granularities, showcasing its stronger parameter utilization capability and better scalability.

\subsection{Robustness of Expert Rank and Expert Count}
We evaluate the performance of TalkLoRA under varying configurations. Specifically, we explore combinations of three per-expert ranks $r_e \in \{8, 16, 32\}$ (where $r_e = r/n$) and three expert counts $n \in \{2, 3, 4\}$, and fine-tune LLaMA3-8B on commonsense reasoning tasks. 

As illustrated in Figure \ref{fig:robustness}, TalkLoRA consistently outperforms existing approaches across all tested configurations. At the lowest rank $r_e=8$, it already attains 86.8\% to 87.6\% accuracy depending on the number of experts. The strongest configuration (4 experts, $r_e=8$) reaches 87.6\%, improving over HiRA by 0.9\% and over standard LoRA by 6.8\%.
Notably, under identical total rank, increasing the number of experts yields higher performance. For instance, at total rank $r=32$, using 4 experts achieves 87.6\%, surpassing 2 experts (87.3\%) by 0.3\%. Similarly, at $r=64$, 4 experts attain 87.4\%, outperforming 2 experts (86.9\%) by 0.5\%.

\subsection{Understanding the TalkLoRA}
Having demonstrated the superiority of TalkLoRA through extensive experiments, we further conduct studies to quantify the contribution of its internal modules and deep understanding of the architecture. Regarding this, we evaluate the importance of the parameter sharing strategy and the Talking Module through the following ablation settings:
1) We remove the combination of layer-unique $E_i$ and shared $B_i$, and instead use unshared $B_i$ matrices.
2) We eliminate the Talking Module, feeding the expert low-dimensional information inputs directly to the router.
The result is shown in Table~\ref{tab:Variant}.

\textbf{Effectiveness Analysis:} TalkLoRA achieves comparable performance to the unshared strategy while using only half the trainable parameters. This substantially reduces redundancy within experts, enabling the MoE mechanism to operate effectively under resource-constrained settings while retaining performance advantages. Removing the Talking Module causes a significant performance drop from 87.6\% to 86.5\%. Despite contributing a negligible number of trainable parameters, the module yields substantial gains.

\textbf{Stability Analysis:}
 Figure \ref{fig:talk} further reveals that the learned $C$ matrices are neither diagonal nor sparse, confirming extensive information exchange across expert. Furthermore, we examine the routing load distribution on the Up matrix for the OBQA dataset in the commonsense reasoning task. Figure \ref{fig:router} clearly shows that models equipped with the Talking Module effectively mitigate overconfident routing. This prevents any single expert from dominating the selection process. The Talking Module significantly improves load balancing, thereby ensuring higher overall model performance.

\section{Conclusion}

In this paper, we propose TalkLoRA, a communication-aware MoELoRA framework that explicitly relaxes the independence assumption among LoRA experts. By introducing a Talking Module, TalkLoRA enables structured information exchange across low-rank experts prior to routing, facilitating coordinated adaptation and more balanced expert utilization. Both theoretical analysis and empirical results show that TalkLoRA enhances the effective expressivity per parameter and improves expert routing, leading to more efficient and robust parameter-efficient adaptation. We believe TalkLoRA offers a principled approach to communication-aware adaptation and demonstrates the potential of structured expert interaction in LLMs.

\section{Limitations}
While we demonstrate improvements on multiple language understanding and reasoning benchmarks, the effectiveness of expert-level communication in broader tasks—such as multi-modal processing, extremely long-context modeling, mathematical reasoning, or code generation—remains to be explored. In addition, due to hardware constraints, we do not evaluate TalkLoRA on very large-scale models. Finally, the incorporation of a MoE-based Talking Module inevitably introduces additional inference latency, which may limit deployment in latency-sensitive applications.

\section{Ethics Statement}
First, as a parameter-efficient fine-tuning (PEFT) technique, TalkLoRA relies on LLMs; consequently, it may inherit or even amplify biases and toxic behaviors present in the base model or the fine-tuning datasets. Users must exercise caution regarding data selection and model evaluation to prevent the propagation of harmful content. Second, the increased accessibility of high-performance fine-tuning on consumer-grade hardware could potentially be exploited by malicious actors to adapt models for generating misinformation or offensive content at a lower cost. We encourage the community to prioritize responsible deployment and incorporate safety alignment measures when utilizing this architecture in real-world applications.

\section{Acknowledgements}
This work is supported by the National Natural Science Foundation of China (No.62206004, No.62572002, No.62272001, No.624065095), and the Natural Science Foundation of Anhui Province (No.2208085QF199, No.2508085MF159, No.2308085MF213).

\bibliography{custom}

\appendix

\clearpage

\onecolumn
\section{Analysis of TalkLoRA}\label{sec:analysis}

We provide a formal analysis supporting the claim that expert-level communication stabilizes routing decisions in TalkLoRA.

\subsection*{Setup}
Let $x \in \mathbb{R}^d$ denote the input to a transformer layer. The low-rank projection is given by $h = Ax$, where $A \in \mathbb{R}^{r \times d}$ is a bounded linear operator. The projected representation is split into $n$ expert subspaces $h = [h_1, \dots, h_n]$. TalkLoRA applies expert communication before routing:
\begin{equation}
    \tilde{h} = (C \otimes I)h,
\end{equation}
where $C \in \mathbb{R}^{n \times n}$ is the expert communication matrix.

The router produces routing probabilities via
\begin{equation}
    g(x) = \text{Softmax}(W_g\tilde{h}),
\end{equation}
with router parameters $W_g$.

\subsection*{Assumptions}
We make the following mild assumptions:
\begin{enumerate}
    \item \textbf{Bounded projection.} \\
    $\|A\|_{\text{op}} \leq \alpha$.
    
    \item \textbf{Non-expansive communication.} \\
    $\|C\|_{\text{op}} \leq 1$.
    
    \item \textbf{Smooth router.} \\
    The router logits are Lipschitz-continuous:
    \begin{equation*}
        \|W_gu - W_gv\| \leq \beta \|u - v\|.
    \end{equation*}
\end{enumerate}
These assumptions are standard and satisfied by common parameterizations used in practice.

\subsection*{Theorem 1 (Routing Stability)}
Under the above assumptions, the routing function $x \mapsto g(x)$ in TalkLoRA is Lipschitz-continuous. Specifically, for any perturbation $\delta x$,
\begin{equation}
    \|g(x + \delta x) - g(x)\| \leq \alpha \beta \|\delta x\|.
\end{equation}

\paragraph{Discussion.}
MoELoRA corresponds to the special case where $C = I$, whereas TalkLoRA allows non-trivial but non-expansive expert communication ($\|C\|_{\text{op}} \leq 1$), which smooths perturbations across experts and leads to more stable routing decisions.

\subsection*{Proof}
A perturbation $\delta x$ propagates through the low-rank projection and expert split, yielding
\begin{equation}
    \|\delta h\| \leq \|A\|_{\text{op}} \|\delta x\| \leq \alpha \|\delta x\|.
\end{equation}
Applying expert communication,
\begin{equation}
    \|\delta \tilde{h}\| = \|(C \otimes I)\delta h\| \leq \|C\|_{\text{op}} \|\delta h\| \leq \|\delta h\|.
\end{equation}
By the Lipschitz continuity of the router logits,
\begin{equation}
    \|W_g\tilde{h}(x + \delta x) - W_g\tilde{h}(x)\| \leq \beta \|\delta \tilde{h}\|.
\end{equation}
Finally, since the Softmax function is Lipschitz on bounded domains, the same bound applies to the routing probabilities. Combining the above inequalities yields the stated result.

\twocolumn
\section{Experimental Setting}
\subsection{Dataset}
\textbf{Commonsense Reasoning:} comprising a total of 170,420 question-answer pairs and 120 random entries as the validation set, and consist of 8 benchmarks and the details are described as follows: \label{dataset:CR}
\begin{itemize}
 \item{BoolQ} \cite{boolq}: This dataset comprises a collection of yes/no question examples, totaling 15942 examples. These questions are naturally occurring and generated in unprompted and unconstrained settings;
 \item {PIQA} \cite{piqa}: This dataset consists of questions with two solutions that require physical commonsense to answer;
 \item {SIQA} \cite{siqa}: This dataset focuses on analyzing people's actions and their social implications;
 \item {HellaS.} \cite{hellaswag}: This dataset consists of commonsense Natural Language Inference (NLI) questions, each featuring a context and multiple endings that complete the context;
 \item {WinoG.} \cite{wino}: This dataset presents a fill-in-a-blank task with binary options. The goal is to select the appropriate option for a given sentence that requires commonsense reasoning;
 \item{ARC-c and ARC-e} \cite{arc}: These two datasets are the Challenge Set and Easy Set of ARC dataset, which contains genuine grade-school level, multiple-choice science questions; 
 \item{OBQA} \cite{obqa}: This dataset comprises questions that require multi-step reasoning, the use of additional common sense knowledge, and thorough text comprehension.
\end{itemize}

\textbf{GLUE:}\label{dataset:glue} benchmark \cite{glue} consists of multiple tasks that target different aspects of natural language understanding, and this study adopts six commonly used datasets among them.
\begin{itemize}
    \item{SST-2} focuses on sentence-level sentiment classification and is evaluated using Accuracy.
    \item{MRP}C is a paraphrase detection task that measures whether a model can determine whether two sentences are semantically equivalent, with Accuracy as its evaluation metric. 
    \item{CoLA} addresses grammatical acceptability judgment, requiring the model to determine whether a sentence is linguistically acceptable, and is evaluated using the Matthews Correlation Coefficient (MCC).
    \item{QNLI} is a natural language inference task converted from a question-answering setting, where the goal is to judge whether a sentence contains evidence that answers the question, and is evaluated using Accuracy. 
    \item{RTE} focuses on recognizing textual entailment, determining whether a premise supports a hypothesis, and is likewise evaluated using Accuracy. 
    \item{STS-B} is designed for semantic similarity assessment, where the model outputs a similarity score from 0 to 5, and its evaluation metrics include Pearson and Spearman correlation coefficients. 
\end{itemize}

Together, these tasks cover essential language understanding abilities, including sentiment analysis, syntactic judgment, semantic similarity modeling, and natural language inference. 
As specified by \cite{RED}, for each benchmark task, we split the public validation set into two parts, as detailed in Table \ref{GLUE}.

\begin{table}[t]
  \centering
    \setlength{\tabcolsep}{1pt}
    \resizebox{\linewidth}{!}{
  \begin{tabular}{lcccccc}
    \hline
    \textbf{Splits Sizes} &\textbf{SST-2} &\textbf{MRPC} &\textbf{CoLA} &\textbf{QNLI} &\textbf{RTE} &\textbf{STS-B} \\
    \hline
    Training Set &67K &3.7K &8.5K &105K &2.5K &5.7K \\
    New Validation Set &436 &204 &522 &1K &139 &750 \\
    New Test Set &436 &204 &521 &4.5K &138 &750 \\
    \hline
  \end{tabular}}
  \caption{\label{GLUE}GLUE dataset sizes.
  }
\end{table}

\begin{table}[t]
  \centering
    \setlength{\tabcolsep}{8pt}
    \resizebox{\linewidth}{!}{
  \begin{tabular}{cccccc}
    \hline
    \textbf{HyperParmaters}   &\textbf{Qwen2.5-7B} & \multicolumn{2}{c}{\textbf{LLaMA2-7B}} & \multicolumn{2}{c}{\textbf{LLaMA3-8B}}  \\
    \hline
    Rank r       & 16 &  16  &  32  & 16   & 32             \\
    $\alpha$     & 16 &  16  &  32  & 16   & 32             \\
    Dropout      &  \multicolumn{5}{c}{0.05}           \\
    Optimizer    &     \multicolumn{5}{c}{AdamW}       \\
    LR           & {1e-4}&   \multicolumn{4}{c}{3e-4}          \\
    LR Scheduler &     \multicolumn{5}{c}{Linear}      \\
    Batch Size   &       \multicolumn{5}{c}{32}         \\
    Warmup Steps&     \multicolumn{5}{c}{100}          \\
    Epochs       &      \multicolumn{5}{c}{2}           \\
    Where        & \multicolumn{5}{c}{Q, K, V, Up, Down} \\
    \hline
  \end{tabular}}
  \caption{\label{Hyperparameters:CR}
    The hyperparameters for TalkLoRA on the commonsense reasoning tasks.
  }
\end{table}
    
\begin{table}[t]
  \centering
    \setlength{\tabcolsep}{2pt}
    \resizebox{\linewidth}{!}{
  \begin{tabular}{lcccccc}
    \hline
 \textbf{HyperParmaters} &\textbf{SST-2} &\textbf{MRPC} &\textbf{CoLA} &\textbf{QNLI} &\textbf{RTE} &\textbf{STS-B}\\
    \hline
    Rank r &\multicolumn{6}{c}{16} \\
    $\alpha$ &\multicolumn{6}{c}{32} \\
    Seed    &\multicolumn{6}{c}{42,43,44,45,46} \\
    Optimizer  &\multicolumn{6}{c}{AdamW} \\
    LR Schedule  &\multicolumn{6}{c}{Linear} \\
    Warmup Ratio &\multicolumn{6}{c}{6e-2} \\
    Max Seq.Len. &\multicolumn{6}{c}{512} \\
    \hline
    Epochs          &40   &50   &50   &40   &60   &80\\
    Learning Rate   &2e-3 &2e-4 &5e-4 &1e-3 &7e-4 &3e-4\\
    Batch Size      &128  &32   &32   &32   &32   &32\\
    \hline
  \end{tabular}}
  \caption{\label{Hyperparameters:glue}
    The hyperparameters for TalkLoRA on the Natural Language Understanding.
  }
\end{table}

\begin{figure*}[t]
\centering
  \includegraphics[width=0.9\linewidth]{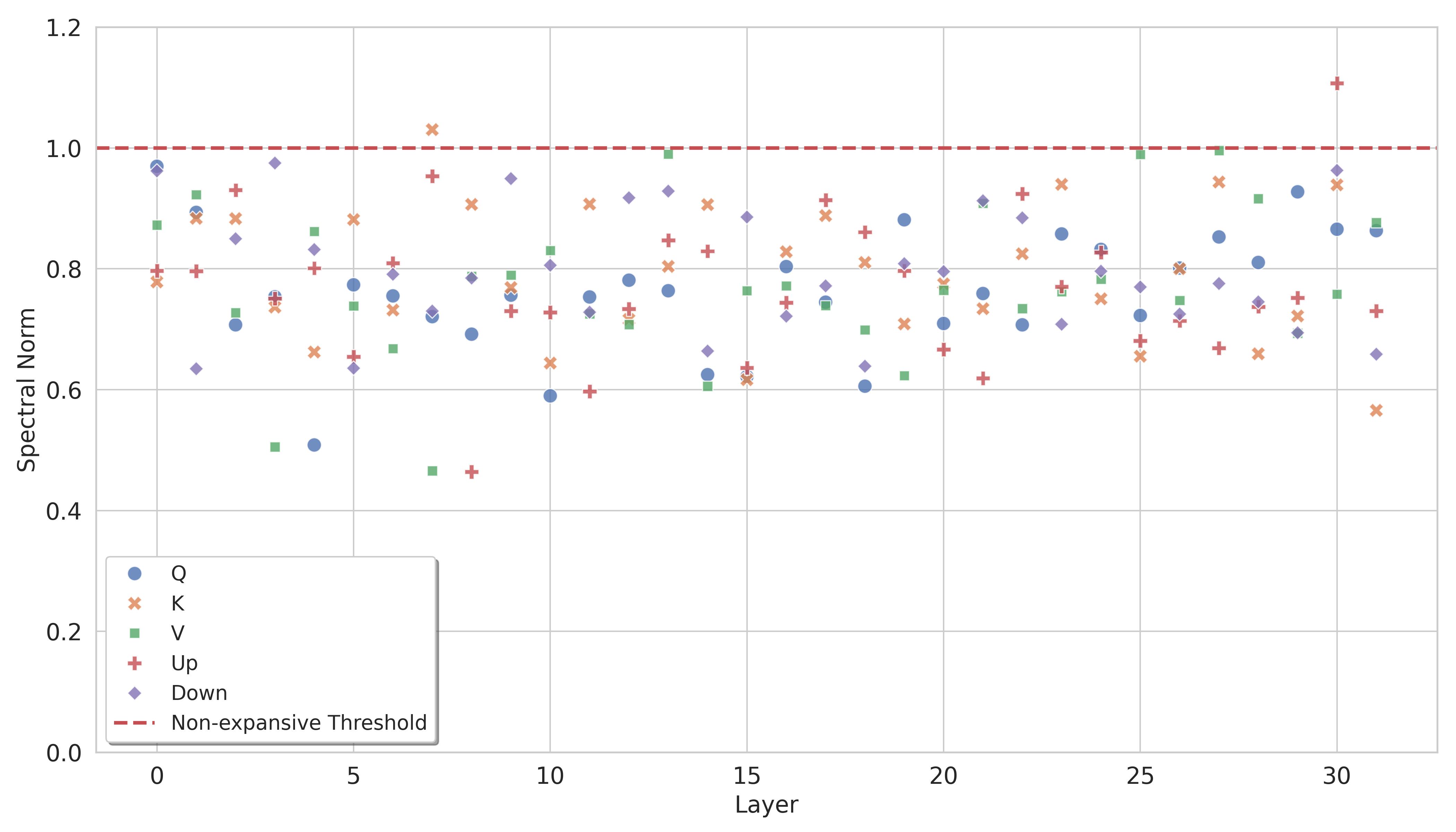} 
  \caption {Distribution of spectral norms for the communication matrix across different layers and modules. The red dashed line ($y=1.0$) denotes the non-expansive threshold. The results demonstrate that the learned matrices predominantly satisfy the non-expansive property, ensuring signal stability throughout the network.
    \label{fig:non-expensive}}
\end{figure*}

\begin{table*}
\centering
    \setlength{\tabcolsep}{10pt}
    \resizebox{\linewidth}{!}{
    \begin{tabular}{ccccccccccc}
    \hline
    \textbf{\#Param(\%)}  & \textbf{TalkLoRA} & \textbf{BoolQ} & \textbf{PIQA} &  \textbf{SIQA} & \textbf{ARC-c} & \textbf{ARC-e} & \textbf{OBQA} & \textbf{HellaS.} &  \textbf{WinoG.}& \textbf{Avg.} \\ \hline
    0.17    & QKV    &74.9 &89.4 &81.2 &83.3 &93.4 &89.4 &96.2 &88.4     & 87.0     \\
    0.24    & UD     &74.0 &90.2 &82.7 &83.2 &93.1 &89.4 &96.4 &89.2     & 87.3     \\
    \hline
    \end{tabular}}
    \caption{Accuracy(\%) comparison of several different tuning granularity of TalkLoRA fine-tuning LLaMA3-8B. Each module is represented by its first letter as follows: (Q)uery, (K)ey, (V)alue, (U)p, (D)own.}
    \label{tab:detailgran}
\end{table*}

\begin{table*}
    \centering
    \setlength{\tabcolsep}{10pt}
    \resizebox{\linewidth}{!}{
    \begin{tabular}{cccccccccccc}
    \hline
    \textbf{\#Param(\%)}  & \textbf{$r_e$} & \textbf{$n$} & \textbf{BoolQ} & \textbf{PIQA} &  \textbf{SIQA} & \textbf{ARC-c} & \textbf{ARC-e} & \textbf{OBQA} & \textbf{HellaS.} &  \textbf{WinoG.}& \textbf{Avg.} \\ \hline
    0.2  &8 &2 &76.1 &88.4 &82.6 &81.5 &93.7 &88.0 &96.0 &88.5     & 86.8     \\
    0.3  &8 &3 &75.7 &89.0 &82.4 &82.6 &93.4 &89.6 &95.9 &89.0     & 87.2     \\
    0.4  &8 &4 &76.1 &89.6 &82.4 &84.5 &93.9 &89.4 &96.0 &89.2     & 87.6     \\
    0.4  &16 &2 &75.6 &89.9 &82.5 &83.4 &93.7 &88.8 &96.1 &88.2     & 87.3     \\
    0.6  &16 &3 &74.7 &88.5 &81.5 &82.7 &94.0 &89.0 &95.8 &89.0     & 86.9     \\
    0.8  &16 &4 &76.3 &89.2 &82.5 &83.5 &93.6 &88.8 &95.8 &89.7     & 87.4     \\
    0.8  &32 &2 &74.3 &88.3 &82.0 &84.0 &93.2 &90.0 &95.7 &87.6     & 87.0     \\
    1.2  &32 &3 &74.2 &88.4 &82.3 &81.6 &91.7 &86.0 &94.4 &87.5     & 85.8     \\
    1.6  &32 &4 &74.2 &88.3 &82.7 &84.4 &93.5 &88.6 &95.9 &89.8     & 87.2     \\
    \hline
    \end{tabular}}
    \caption{Accuracy(\%) comparison of robustness for TalkLoRA fine-tuning LLaMA3-8B.}
    \label{tab:robustness}
\end{table*}

\begin{table*}[!t]
    \centering
    \setlength{\tabcolsep}{10pt}
    \resizebox{\linewidth}{!}{
    \begin{tabular}{ccccccccccc}
    \hline
    \textbf{\#Param(\%)}  & \textbf{Method} & \textbf{BoolQ} & \textbf{PIQA} &  \textbf{SIQA} & \textbf{ARC-c} & \textbf{ARC-e} & \textbf{OBQA} & \textbf{HellaS.} &  \textbf{WinoG.}& \textbf{Avg.} \\ \hline
    0.7   &w/o Sharing  &75.4 &89.8 &83.0 &84.5 &93.7 &89.4 &96.0 &89.8     & 87.7     \\
    0.4   &w/o Talking  &72.6 &88.8 &82.2 &83.4 &93.0 &88.4 &95.7 &88.2     & 86.5     \\
    \hline
    \end{tabular}}
    \caption{Accuracy(\%) comparison of variant of TalkLoRA fine-tuning LLaMA3-8B.}
    \label{tab:detail_variant}
\end{table*}

\subsection{Hyperparameters}
Table~\ref{Hyperparameters:CR} shows the detailed hyperparameters for commonsense reasoning tasking when fine-tuning the Qwen2.5-7B, LLaMA2-7B and LLaMA3-8B. Table~\ref{Hyperparameters:glue} shows the detailed hyperparameters for GLUE benchmark when fine-tuning the RoBERTa-base.

\section{Additional Experimental} \label{sec:add experimental}
\subsection{Non-expansive validation}
As discussed in the main text, a non-expansive communication matrix $C$ ensures that perturbations in the input are not amplified during transmission. This property guarantees bounded changes in the router's input, which is critical for the numerical stability of the Mixture-of-Experts (MoE) architecture during inference. To empirically verify this property in TalkLoRA, we extract the learned communication matrices from the Talking Module across all layers of the fine-tuned model. For each communication matrix $C \in \mathbb{R}^{n \times n}$, we computed its spectral norm, defined as the largest singular value of the matrix:
\begin{equation}
    \|C\|_2 = \sigma_{\max}(C) = \max_{x \neq 0} \frac{\|Cx\|_2}{\|x\|_2}.
\end{equation}
A matrix is non-expansive in the Euclidean space if its spectral norm satisfies $\|C\|_2 \le 1$. We evaluate this quantity for all projection types (Q, K, V, Up, and Down) across all transformer layers. The resulting distributions of spectral norms are shown in Figure~\ref{fig:non-expensive}. where the x-axis denotes the layer index and the y-axis denotes the corresponding spectral norm. We observe that the spectral norms of the learned communication matrices consistently remain at or below 1 across layers and projection types. These results provide empirical evidence that the Talking Module in TalkLoRA learns non-expansive communication operators in practice, supporting the stability assumptions underlying our theoretical analysis of routing robustness.
\subsection{Tuning Granularity Analysis}
This section is a detail experiment result of adapting different weight modules using TalkLoRA. Each module is represented by its first letter as follows: (Q)uery, (K)ey, (V)alue, (U)p, (D)own. We conduct experiments using LLaMA3-8B with a rank of 32 and expert count of 4 on commonsense reasoning training samples. The result, shown Table~\ref{tab:detailgran}

\subsection{Robustness of TalkLoRA}
This section presents detailed experimental results examining the robustness of TalkLoRA. We explore combinations of three per-expert ranks $r_e \in \{8, 16, 32\}$ (where $r_e = r/n$) and three expert counts $n \in \{2, 3, 4\}$, and fine-tune LLaMA3-8B on commonsense reasoning tasks. The result is shown in Table~\ref{tab:robustness}

\subsection{Ablation Study of TalkLoRA}
This section presents detailed experimental results investigating the two variants of TalkLoRA. We conduct experiments using LLaMA3-8B with a rank of 32 and expert count of 4 on commonsense reasoning training samples. The result is shown in Table~\ref{tab:detail_variant}.

\end{document}